%% file: main.tex
\documentclass{article}
\usepackage[utf8]{inputenc}

\usepackage{amsfonts}

\usepackage[dvipsnames]{xcolor}

\usepackage{geometry}
\usepackage[citestyle=authoryear, backend=biber, sorting=ynt, natbib=true, url=false, doi=false]{biblatex}
\newcommand{\Call}[2]{\text{\textsc{#1}(#2)}}

\usepackage{amsmath}

\usepackage[utf8]{inputenc}
\usepackage[T1]{fontenc}
\usepackage{mathpazo} 

\usepackage{booktabs}

\usepackage[parfill]{parskip} 

\usepackage{authblk}

\usepackage{hyperref}

\usepackage{pgf, pgfplots, tikz}
\pgfplotsset{compat=newest}
\usepackage{subcaption}
\usepgfplotslibrary{groupplots}
\usepgfplotslibrary{statistics}

\title{Guidelines for the Computational Testing of Machine Learning approaches to Vehicle Routing Problems}

\author[1]{Luca Accorsi}
\author[2, 3]{Andrea Lodi}
\author[1, 4]{Daniele Vigo}

\date{\nolinkurl{{luca.accorsi4, daniele.vigo}@unibo.it}\\
      \nolinkurl{andrea.lodi@polymtl.ca}}

\affil[1]{Department of Electrical, Electronic and Information Engineering ``G. Marconi'', University of Bologna, Italy}
\affil[2]{Canada Excellence Research Chair in Data Science for Decision Making, {\'E}cole Polytechnique de Montr{\'e}al, Canada}
\affil[3]{Mila, Quebec Artificial Intelligence Institute, Canada}
\affil[4]{CIRI ICT, University of Bologna, Italy}

\begin{document}

\maketitle

\begin{abstract}
    \noindent
    Despite the extensive research efforts and the remarkable results obtained on Vehicle Routing Problems (VRP) by using algorithms proposed by the Machine Learning community that are partially or entirely based on data-driven analysis, most of these approaches are still seldom employed by the Operations Research (OR) community. 
    Among the possible causes, we believe, the different approach to the computational evaluation of the proposed methods may play a major role.
    With the current work, we want to highlight a number of challenges (and possible ways to handle them) arising during the computational studies of heuristic approaches to VRPs that, if appropriately addressed, may produce a computational study having the characteristics of those presented in OR papers, thus hopefully promoting the collaboration between the two communities.
\end{abstract}

\section{Introduction} \label{sec:introduction}
Recently, several attempts to use machine learning (ML) to deal with combinatorial optimization problems (COPs) have been proposed. 
Indeed, ML promises to support the design of algorithms as well as the resolution process, by decreasing the need for hand-crafted and specialized solution approaches, which are notably known to require high expertise and a huge time investment for being developed.
The incredible advances in deep learning (DP, \citet{Goodfellow-et-al-2016}) coupled with increasingly powerful hardware, have led to very promising results for a variety of COPs (\citet{bengio2020machine, mazyavkina2020reinforcement}).

In this work, we focus on COPs arising in the area of vehicle routing. Vehicle routing problems (VRPs, \citet{VRPBOOK}) are, in fact, increasingly receiving attention in the ML community both because of their relevance in real-world applications and the computational challenges they pose (\citet{Vesselinova_2020}).
Indeed, VRPs have been the test-bed for novel neural network architectures such as pointer networks (\citet{NIPS2015_29921001}), graph embedding networks (\citet{dai2016discriminative, Khalil2017learning}) and attention-based models (\citet{kool2019attention, deudon2018learning}) achieving stunning results on a variety of problems.
Currently, most of the proposed approaches aim at learning constructive heuristics, which sequentially extend a partial solution, possibly employing additional procedures such as sampling and beam search (see e.g., \citet{bello2017neural} and \citet{hottung2021efficient}). Few others, such as \citet{wu2020learning} and \citet{NEURIPS2019_131f383b}, instead, focus on learning improvement heuristics to guide the exploration of the search space and iteratively refine an existing solution.

Despite of the relevance of the problems, the increasing attention they are receiving in the ML community and the remarkable results achieved so far, these techniques have not yet become widespread in the Operations Research (OR) community. 
This may be because of their novelty and the nontrivial effort required to adopt them by the OR community that typically has a different background.
But also, at the same time, the computational testing provided by some of the proposed ML methods may not be very convincing with respect to the standard practices in the OR and VRP community. 
The current work focuses on this second aspect by highlighting important points to consider during the computational testing and providing guidelines and methodologies that we believe are relevant from an OR perspective. 
The remainder of the paper is organized as follows. Section \ref{sec:guidelines} surveys crucial aspects faced during a computational testing such as the selection of proper benchmark instances and baseline algorithms, as well as techniques to properly compare different solution approaches. Section \ref{sec:examples} provides concrete examples and pointers to representative computational studies found in recent papers. Finally, concluding remarks are given in Section \ref{sec:conclusions}.

\section{Guidelines for a Computational Testing} \label{sec:guidelines}
In this section, we highlight important points an OR practitioner would consider crucial when examining the computational results section of a novel approach.

\subsection{Benchmark Instances and Problem Definition}
The training phase of a ML-based approach typically requires a large number of VRP instances sharing the same characteristics. 
A common way to generate these instances is by randomly defining their characteristics, sampled from arbitrarily defined distributions. 
In the following, we argue that adopting a common problem description as well as a common set of benchmark instances is extremely important to correctly assess the potentialities of a novel approach.

\textit{Problem description and objectives.} 
The term VRP identifies a class of problems containing an enormous number of different variants. Among the most studied ones, we have the Capacitated Vehicle Routing Problem (CVRP), the Vehicle Routing Problem with Time Windows (VRPTW), and the Vehicle Routing Problem with Pickup and Deliveries.
In the following, if not specified differently, we limit our treatment to the CVRP which is often taken as a reference problem in ML-based approaches and, despite its simple definition, is still a challenging problem for both traditional and innovative approaches.
We refer to Chapter 1 of \citet{VRPBOOK} for a thorough overview of VRPs. Specific VRPs have widely accepted formulations and precise problem definitions. For example, modern CVRP instances do not fix a-priori the number of routes a solution should have or heuristic approaches to the VRPTW typically consider a hierarchical objective: first minimize the number of vehicles and then the routing cost. It is thus clear that the specific definition of the problem has a crucial impact on the obtained results.

\textit{Test instances representativeness.} Despite VRPs being $\mathcal{NP}$-hard, the actual challenge posed by a specific instance is highly dependant on several factors such as customers and depot location, the vehicle capacity and the customers demand distribution.
The VRP community has thus identified, for each specific problem in the VRP class, a set of benchmark instances that is currently considered to be relevant for testing modern approaches.
Thus, in addition to possible instances defined by the ML community, we highlight the importance of using, whenever possible, instances derived from these widely recognized, used and studied datasets. 

As an example, in \citet{UCHOA2017845}, the authors introduced the $\mathbb{X}$ instances for the CVRP, thoroughly describing their generation process. 
This process can then be emulated to generate (possibly smaller-sized) more representative CVRP instances, as it was done in \citet{kool2021deep} and in \citet{hottung2021efficient} (appendix). In \citet{wu2020learning}, the authors directly used a subset of $\mathbb{X}$ instances along with distributions more commonly used in other ML works.

As an additional example, consider the VRPTW for which the so-called Solomon instances and their extension proposed by \citet{solomon1987} and \citet{gehring1999parallel}, respectively, are the current benchmark. Also for them, \citet{solomon1987} describes the procedure used to define the time window constraints that can thus be considered when defining new instances.

\textit{Repositories of instances.} 
Together with papers introducing or using certain sets of instances (whose authors could be contacted to retrieve), we mention two among the most popular repositories of VRP instances. Namely, \href{http://www.vrp-rep.org/}{\citeauthor{vrprep}} collects instances for more than 50 different VRP variants, best known solution values, and references to papers obtaining these results. In addition, \href{http://vrp.atd-lab.inf.puc-rio.br/index.php/en/}{\citeauthor{CVRPLIB}} contains instances and up-to-date best-known solutions of CVRP instances.
Finally, small instances commonly used in the past or in exact methods can be found in \href{http://comopt.ifi.uni-heidelberg.de/software/TSPLIB95/}{\citeauthor{tsplib}} and \href{http://people.brunel.ac.uk/~mastjjb/jeb/info.html}{\citeauthor{orlib}}.

\subsection{Baseline Algorithms}
The selection of a proper baseline algorithm is extremely important, failing in this task would hinder the objective evaluation of the potential of a novel approach. 
Indeed, since results (computing time, solution quality, and possibly more sophisticated measures as detailed in Section \ref{sec:comparison}) are crucial to compare different solution approaches over a common set of problems and instances, a wrong baseline may distort their interpretation, undermining the whole validation process. Despite the purpose of ML-based approaches not being that of outperforming highly specialized solvers, but rather that of proposing versatile tools not requiring high-levels of manual engineering, the comparison should still occur against the best performing algorithms to better comprehend the tradeoff between data-driven and ad-hoc algorithms.

\textit{Include the best available algorithms.}
Along with simple baselines and competing ML-based methods, we argue that one should consider the inclusion of the best available algorithms proposed by the OR community for each specific VRP.

The selection of baseline algorithms is often guided by the availability of free-to-use or open-source reliable software packages.
Fortunately, more and more researchers are publishing the source code of heuristic as well as exact state-of-the-art VRP solvers that can be freely used for research activities.

\paragraph{Heuristic solvers}
The widely used Google OR-Tools (\citet{ortools}) is erroneously considered by most ML papers to be among the best open-source VRP solvers (see, e.g., \citet{nazari2018reinforcement}) while achieving on the CVRP far-from-optimal results on the $\mathbb{X}$ instances (see \citet{vidal2020hybrid}). Much better open-source solvers for the CVRP are fortunately available. Along with LKH-3 (\citet{lkh3}), which is already widely used by the ML community for solving VRPs, we mention HGS-CVRP (\citet{vidal2020hybrid}) and FILO (\citet{filo}) as highly effective and efficient open-source heuristic solvers for the CVRP that, on the widely studied $\mathbb{X}$ instances of the CVRP (\citet{UCHOA2017845}), produce superior results compared with LKH-3 (see \citet{lscgh}). Finally, we mention SISR (\citet{ChristiaensJanSIbS}), that, despite not being already available in terms of source code, is conceptually simple and easy to implement, yet providing state-of-the-art results on a great variety of VRPs.

\paragraph{Exact solvers}
Several papers solve small instances to optimality by using general-purpose optimization solvers such as Gurobi or CPLEX. Despite the noble attempt, trying to directly solve simple compact VRP formulations by using a generic branch-and-cut approach would soon turn out to be an extremely challenging task.
Indeed, several papers report the ability of solving only very small instances (e.g., CVRPs with about 20 customers). 
Instead, VRPSolver (\citet{Pessoa2020}), a freely available (for academic purposes) exact solver specialized for routing problems, should be considered for serious and reliable testing of VRPs. Indeed, VRPSolver combines a branch-cut-and-price algorithm with other sophisticated techniques specifically designed for VRPs such as route enumeration, state space relaxation, and others being able to consistently solve CVRP instances with up to 200 customers (a size already out of reach for most ML-based approaches proposed so far).

\subsection{Algorithms Comparison} \label{sec:comparison}
Comparing algorithms having a completely different nature is an extremely challenging task. On the one hand, traditional solution approaches proposed by the OR community are designed to handle set of instances having a broad range of different characteristics. Moreover, these approaches are typically tuned to achieve, with a single set of parameter values, results that are on average good over all tested instances.
On the other hand, ML-based approaches generally need to treat every instance distribution separately, requiring a specific tuning and thus additional training (possibly taking up to several weeks of computing time).
In addition, traditional OR algorithms are (almost) always executed on CPUs using a single thread, while ML-based approaches naturally benefit from running on massively parallel hardware architectures such as GPUs. Finally, the programming language may also play a role. In fact, traditional OR algorithms are usually implemented in highly efficient languages such as C++, whereas ML-based approaches typically use Python that mixes slow interpreted code with efficient native libraries.

\textit{Facilitate comparisons (i.e., run all solvers by using a common configuration).} In production, algorithms should obviously make fully use of the best existing available technologies. 
In fact, even traditional algorithms may contain (portions of) embarrassingly parallel code that would thus benefit from being run on multiple threads. 
As an example, a common dynamic programming procedure employed in VRP exact solvers would greatly benefit from a GPU implementation compared to a traditional sequential version (\citet{boschetti2017route} report speedup of up to 40 times). Another well-known example is the parallel implementation of Branch \& Bound algorithms (see, e.g., \citet{parallelbb2006}).
Despite the clear time-savings of a parallel implementation, experimental evaluation asks to reduce at a minimum the different factors that would render comparisons among approaches unnecessarily more challenging. It is thus commonly accepted to consider single-threaded algorithms run on standard CPU architectures. 
A very interesting information, that would promote a direct comparison of newly proposed algorithms with existing ones, consists in including also the computing time taken by the algorithm when run with the above defined settings.
If running all the experiments both on GPU and on CPU with a single thread would be computationally prohibitive, we argue that one should consider adding a measure of the speedup associated with the model inference when run on a GPU rather than on a CPU together with the total algorithmic time (in which everything except the model inference is run on a CPU with a single thread), and the fraction of time spent doing inference on GPU. This way, the total running time of the algorithm on a CPU with a single thread could be easily estimated. Another possibility sees the inclusion of a rough measure of the number of CPU cores needed to match the GPU capacity as it was done in \citet{hottung2021efficient}.


\textit{Convert computing time to a common scale.} When using results published on other papers, because the source code may not be available or running again the experiments may be too time consuming, a common practice consists in roughly scaling the computing time to a common base by using appropriate factors. 
A practice increasingly adopted in the VRP community consists in using the single-thread rating defined by \citeauthor{passmark}. 
So that, given a base processor, say an Intel \href{https://www.cpubenchmark.net/cpu.php?cpu=Intel+Xeon+E3-1245+v5+\%40+3.50GHz&id=2674}{Xeon CPU E3-1245 v5} having a single-thread rating of 2277, and a target processor, say an \href{https://www.cpubenchmark.net/cpu.php?cpu=Intel+Core2+Duo+T5500+\%40+1.66GHz&id=922}{Intel Core2 Duo T5500} having a single-thread rating of 594, the computing time of an algorithm run on the target processor is reduced of a factor $\approx 3.83$.
The comparison is still rough, being the CPU just one among the several components affecting the overall performance. However, this is one of the possible approaches generally accepted as sound by the VRP community.
Note that, the above considerations assume all algorithms have been run on an unloaded system and on a CPU by using a single thread.

\textit{On the comparison with exact solvers.} Exact solvers are often used as baseline algorithms when approaching small-to-medium size instances. When reporting the computing time spent by an exact solver, especially if its results are compared with a heuristic algorithm, it should be considered that the former may find very good quality solutions early during the run and then spend the majority of the time to prove optimality. 
Thus, when reporting results obtained by an exact solver, an additional column could be added showing the computing time in which the last solution (i.e., the optimal one, if the solver is not prematurely stopped) was found. 
Despite being true that the solver does not have a termination criterion to know whether the solution at hand is optimal or not, being compared with heuristics, this approach would still provide a feeling on the convergence speed of the solver.
Another possibility, which however requires a much more detailed collection and processing of data, consists in realizing a convergence profile chart (see later). We finally mention the primal integral introduced by \citet{Achterberg2012}, as a measure able to take into account the overall solution process in terms of convergence towards the optimal (or best known) solution over the entire solving time.

\textit{Consider the statistical relevance of the results.}
In the past, it was common practice to use just average percentage errors with respect to the best known solution to compare the performance of different algorithms. However, more recently, the inclusion of simple statistical tests to objectively assess the differences among algorithms is increasingly becoming popular in the VRP community.
A common practice consists in using a one-tailed Wilcoxon signed-rank test (see \citet{wilcoxon}) possibly coupled with correction methods when multiple comparisons involving the same data are performed (e.g., the Bonferroni correction, see \citet{dunn}).
These tests are used to determine whether two sets of paired observations, for which no assumption can be done on their distribution, are statistically different, and thus, whether two algorithms are considered to provide equivalent results. 
A common tool for performing these statistical tests is the R language (\citeauthor{rcore}), which allows to execute them with just a few lines of code.
An example of a test application can be found in Section \ref{sec:stattest}. 

\textit{Always compare averages.} 
When experimenting with algorithms containing randomized components, we argue that one should make comparison by considering the average results obtained over a reasonable number of runs. Typically, the used number of runs are 10 or 50, that, despite not being statistically significant, allow to qualitatively identify whether an algorithm provides stable or highly variable results.

\textit{Include charts.}
Charts allow to visually compare several algorithms at once, thus providing a fast and effective way to view the results, that, when coupled with tables and statistical tests, give a satisfying and thorough picture of the computational studies for a set of instances.
\begin{figure}
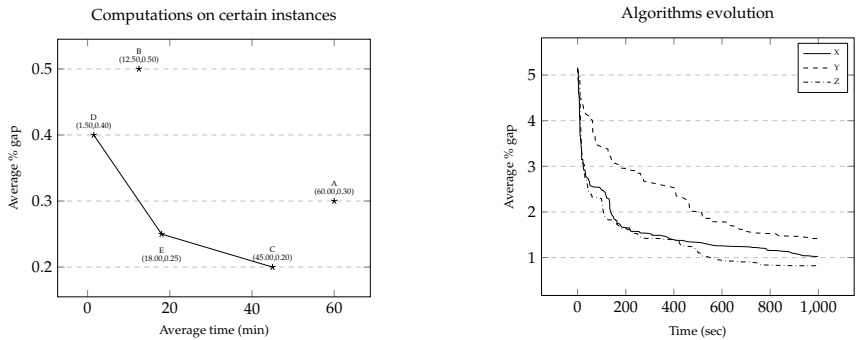

\centering
  \begin{subfigure}[b]{0.4\textwidth}
    \include{teximgs/performance-chart}
  \end{subfigure}
  \begin{subfigure}[b]{0.4\textwidth}
    \include{teximgs/convergence-profile-chart}
  \end{subfigure}
\caption{On the left side, the performance chart showing for  algorithms A, B, C, D, and E the average normalized time $t$ and solution quality obtained in a number $n$ of runs with different seeds. Algorithms C, D, and E dominate A and B. On the right side, the search trajectory defined by the average gap found at a given time instant for algorithms X, Y and Z.}
\label{fig:charts}
\end{figure}
Among the most used charts, we mention
\begin{itemize}
    \item the performance chart, relating the average normalized computing time with the average solution quality (e.g., the percentage gap with respect to a reference value, which is typically that of the optimal or best known solution) of each algorithm in the comparison, see Figure \ref{fig:charts} (left);
    \item the convergence profile chart, showing for each time instant the average solution quality for the compared algorithms, see Figure \ref{fig:charts} (right).
\end{itemize}
The performance chart clearly identifies Pareto optimal algorithms and dominated ones, whereas the convergence profile chart shows their converge speed when executed for a specific period of time. Convergence profile charts can be used for improvement heuristics but also for simpler constructive heuristics provided that they include additional iterative procedures used to further improve the solution quality (e.g., the active search of \citet{bello2017neural}).

\textit{On the choice of the programming language.} (Baseline) Algorithms should be implemented efficiently. This may include using appropriate data structures as well as programming languages producing directly executable machine code. An efficient implementation of a competing algorithm should not be considered negatively.

\textit{Consider the analysis of the various algorithm components.}
Several papers include additional analysis on the components of the proposed algorithm. This may include the behavior of the algorithm when some parameters are changed as well as the contribution of individual components to the overall final results (e.g., the average improvement of different local search operators analyzed throughout the algorithm execution). 
This kind of analysis is both considerably appreciated in the VRP community and extremely important to grasp insights on the overall contribution and usefulness of the different components of an algorithm.

\section{Examples of a Computational Studies} \label{sec:examples}
To make the above suggestions more concrete, in this section, we report extracts of and pointers to representative computational studies found in recent papers on the CVRP. 
We will consider two scenarios: in the first one, we assume to have all source codes available, whereas in the second one, we assume at least one of the competing algorithm source code is not available to the tester.

\subsection{Scenario 1: all algorithms source codes are available}
This is the simplest and most fortunate setting. Indeed, given enough time and processing power, all algorithms can be run on exactly the same platform and for the same amount of time. In this context the convergence profile chart perfectly shows the evolution of each algorithm, making evident its speed to reach certain quality levels. 

Table \ref{table:scenario1} reports a rearranged extract of Tables 1-3 proposed in \citet{vidal2020hybrid} comparing three of the above mentioned algorithms (more specifically HGS-CVRP, SISR, and OR-Tools) run on a common platform for the same computing time and over the same $\mathbb{X}$ instances of the CVRP. In particular, the table shows for each competing algorithm the average solution value (Avg) and the associated average gap (Gap) computed considering a certain number of runs of the algorithm when it includes randomized components. The gap is computed with respect to a reference best known solution value (BKS) as Gap = $100 \cdot (\textrm{Avg} - \textrm{BKS}) / \textrm{BKS}$. Additional useful columns could include the worst and best gap obtained, so as to better examine the variability of the quality for the algorithm on the dataset.
\begin{table}
    \footnotesize
    \centering
    \begin{tabular}{lrrrrrrrrrr}
\toprule
&&& \multicolumn{2}{c}{HGS-CVRP} && \multicolumn{2}{c}{SISR} && \multicolumn{2}{c}{OR-Tools} \\
\cmidrule{4-5}
\cmidrule{7-8}
\cmidrule{10-11}
Instance & BKS && Avg & Gap && Avg & Gap && Avg & Gap \\
\toprule
X-n101-k25 & 27591      && 27591.0 & 0.00       && 27593.3 & 0.01       && 27977.2 & 1.40\\
X-n106-k14 & 26362      && 26381.4 & 0.07       && 26380.9 & 0.07       && 26757.5 & 1.50\\
X-n110-k13 & 14971      && 14971.9 & 0.00       && 14972.1 & 0.01       && 15099.8 & 0.86\\
\multicolumn{1}{c}{\ldots}\\
X-n979-k58 & 118987     && 119247.5 & 0.22      && 119108.2 & 0.10      && 123885.2 & 4.12\\
X-n1001-k43 & 72359     && 72748 & 0.54       && 72533.1 & 0.24       && 78084.7 & 7.91\\
\midrule
Mean &&& & 0.11 && & 0.19 && & 4.01\\ 
\bottomrule
\end{tabular}
    \caption{Minimal table showing the solution quality obtained by algorithms run for the same amount of time on a common platform.}
    \label{table:scenario1}
\end{table}
Finally, we refer to Figures 3 and 4 of \citet{vidal2020hybrid} for examples of convergence profile charts for the above-mentioned algorithms.  

\subsection{Scenario 2: at least one algorithm source code is not available}
This scenario, which is still unfortunately frequent in the VRP community, makes the comparison of results much more challenging. The common case sees only the presence of tables showing the performance of a proposed algorithm over a set of benchmark instances when run on a certain platform. First of all, an assumption is required when reviewing these tables reporting the computational results. In particular, we have to assume that the results published in the paper, which are inevitably obtained with a specific tuning of parameters, are the values on which the authors desire to compare with other approaches. We shall note that these parameters do include the termination criterion, which was then selected as a design choice, and considered to provide competitive results (otherwise a different criterion would have been selected).
Since data on the convergence of the algorithm are typically not available, the comparison can then be performed over the published results by first normalizing the computing time in the best possible way and then analyzing the bi-objective perspective provided by the performance chart shown in \ref{fig:charts} (left) showing non dominated algorithms for a fixed configuration of their parameters.

\subsection{Statistical validation of the results}
\label{sec:stattest}
In both the above scenarios (and provided that the average solution quality obtained by an algorithm is available for each instance in the dataset), the (final) solution quality could be assessed with simple statistical tests. This procedure is useful especially when the proposed methods do not clearly dominate the others, for example because they obtain similar average percentage gaps on the considered benchmark instances.

\begin{figure}[b]
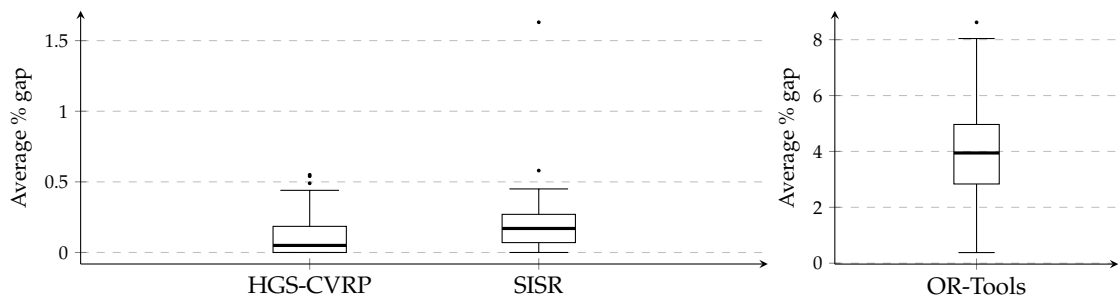

    \centering
    \include{teximgs/boxplot-groups}
    \caption{Average gaps obtained on the $\mathbb{X}$ instances by HGS-CVRP, SISR and OR-Tools. Note the different y-axis for OR-Tools. The thick line identifies the median value.}
    \label{fig:boxplots}
\end{figure}

As an example, in the following we compare HGS-CVRP, SISR, and OR-Tools on the $\mathbb{X}$ instances. Data have been taken from Tables 1 and 3 of \citet{vidal2020hybrid} and is summarized in the boxplots of Figure \ref{fig:boxplots}.

Similarly to what was done in \citet{ChristiaensJanSIbS}, we can assess the results obtained by our proposed algorithm, say HGS-CVRP, against the remaining ones, by conducting a one-tailed Wilcoxon signed-rank test in which we consider a null hypothesis $H_0$
\begin{equation*}
    H_0: \Call{AvgSolCost}{HGS-CVRP} = \Call{AvgSolCost}{X},
\end{equation*}
and an alternative hypothesis $H_1$ 
\begin{equation*}
    H_1: \Call{AvgSolCost}{HGS-CVRP} > \Call{AvgSolCost}{X},  
\end{equation*}
where $X$ can be SISR and OR-Tools. A hypothesis is rejected when its $p$-value is lower than a significance level $\alpha$. In particular, we have that
\begin{itemize}
    \item failing to reject $H_0$ means that the average results of the two methods are not statistically different;
    \item whereas, when $H_0$ is not rejected, the average results are statistically different and the alternative hypothesis $H_1$ can be tested to find whether they are greater than those of a competing method. Rejecting $H_1$ thus implies that HGS-CVRP performs better than the competing method.
\end{itemize}
Moreover, as mentioned in Section \ref{sec:comparison}, when performing multiple comparisons involving the same data, the probability of erroneously rejecting a null hypothesis increases. To control these errors, the significance level $\alpha$ is adjusted to lower values. Bonferroni correction (\citet{dunn}) is a simple method that can be used for this purpose. In particular, given $n$ comparisons, the significance level is set to $\alpha / n$.

In the above comparison we tested a total number of $n=2$ hypothesis corresponding to the partitioning of instances (1, all $\mathbb{X}$ instances together) and to the two hypothesis (2, $H_0$ and $H_1$). Assuming an initial significance level $\alpha_0 = 0.025$, the adjusted value becomes $\alpha = 0.025 / 2 = 0.0125$. 

We can compute the $p$-values, which are shown in Table \ref{tab:pvalues}, for our analysis for example by using the R language.
\begin{table}
    \centering
    \begin{tabular}{l rrr}
    \toprule
    && SISR & OR-Tools \\
    \midrule
    $H_0$ && 8.27934e-06 & 3.95591e-18 \\
    $H_1$ && 4.13967e-06 & 1.97796e-18  \\
    \bottomrule
    \end{tabular}
    \caption{$p$-values obtained when comparing HGS-CVRP with SISR and OR-Tools.}
    \label{tab:pvalues}
\end{table}
For both SISR and OR-Tools we have that the associated $p$-value is lower than the significance level $\alpha$. We can thus reject the hypothesis $H_0$ that the average results of HGS-CVRP are similar to those of SISR and OR-Tools. Finally, by testing $H_1$ we again are able to reject the hypothesis, concluding that HGS-CVRP average results are statistically better than those obtained by SISR and OR-Tools on the $\mathbb{X}$ instances.


\section{Conclusions} \label{sec:conclusions}
With this work we highlighted several challenges arising when comparing traditional and machine learning-based solution approaches for vehicle routing problems. Our aim was that of providing a set of reference guidelines that would help the machine learning community to produce computational studies that would be better appreciated by the operations research, and especially the vehicle routing, community. Finally, we conclude by referring the reader to the excellent overview on experimental analysis by \citet{Johnson1999ATG}.

\printbibliography

\end{document}

%% file: teximgs/performance-chart.tex
	\begin{tikzpicture}[scale=0.60]
	\begin{axis}[
	enlargelimits=0.15,
	enlarge x limits=0.15,
	title={Computations on certain instances},
	ymajorgrids=true,
	grid style=dashed,
	black, fill=black,	
	xlabel={\footnotesize Average time (min)},
	ylabel={\footnotesize Average \% gap},
	]

    \addplot[color=black, solid, label=none ] coordinates {(1.50,0.40)(18.00,0.25)(45.00,0.20)};
    
    \addplot+[color=black, only marks, point meta=explicit symbolic, mark=star, nodes near coords, text width=3cm, align=center, font=\tiny\linespread{0.8}\selectfont] 
    coordinates {
        (60.00,0.30)[\tiny{A\\(60.00,0.30)}]
        (12.50,0.50)[\tiny{B\\(12.50,0.50)}]
        (1.50,0.40)[\tiny{D\\(1.50,0.40)}]
        (45.00,0.20)[\tiny{C\\(45.00,0.20)}]
    };
    
    \addplot+[color=black, only marks, point meta=explicit symbolic, mark=star, nodes near coords, text width=3cm, align=center, font=\tiny\linespread{0.8}\selectfont, every node near coord/.append style={yshift=-0.75cm}] coordinates {
        (18.00,0.25)[\tiny{E\\(18.00,0.25)}]
    };

	\end{axis}
	\end{tikzpicture}

%% file: teximgs/convergence-profile-chart.tex
	\begin{tikzpicture}[scale=0.60]
	\begin{axis}[
	enlargelimits=0.15,
	enlarge x limits=0.15,
	title={Algorithms evolution},
	ymajorgrids=true,
	grid style=dashed,
	black, fill=black,	
	xlabel={\footnotesize Time (sec)},
	ylabel={\footnotesize Average \% gap},
	]

    \addplot[color=black, solid, label=none ] table [x=iter, y=gap, col sep=comma] {csvs/alg1evo.csv};
    
    \addplot[color=black, dashed, label=none ] table [x=iter, y=gap, col sep=comma] {csvs/alg2evo.csv};

    \addplot[color=black, dashdotted, label=none ] table [x=iter, y=gap, col sep=comma] {csvs/alg3evo.csv};

    \legend{\tiny{X},\tiny{Y},\tiny{Z}}

	\end{axis}
	\end{tikzpicture}

%% file: teximgs/boxplot-groups.tex
\scriptsize

\begin{tikzpicture}

\begin{groupplot}[group style={group size=2 by 1, vertical sep=0cm, horizontal sep=25pt}]

\nextgroupplot[
    clip=false,
	boxplot/draw direction=y,
	boxplot/variable width,
	boxplot/every median/.style={black,very thick,solid},
	axis lines=left,
	xmin=0.5,xmax=2,
	width=0.66\textwidth,
	height=0.20\textheight,
	ylabel style={align=center}, 
	y tick label style={align=right},
	x tick label style={align=center},
	xtick={1, 1.5},
	enlarge y limits=0.05,
	ymajorgrids=true,
	grid style=dashed,
	xticklabels={{\small HGS-CVRP}, {\small SISR}}, xticklabel style={rotate=0}, ylabel={\footnotesize Average \% gap},
	scatter/classes={ a={mark=star}, b={mark=*}}]

	\addplot[
	mark=*, mark size=0.5pt,boxplot,
	boxplot prepared={draw position=1, 
	lower whisker=0.00, 
	lower quartile=0.00,  
	median= 0.05, 
	upper quartile=0.185, 
	upper whisker=0.440,sample size=1}
	] coordinates{ (1, 0.55) (1, 0.49) (1, 0.54) };

	\addplot[
	mark=*, mark size=0.5pt,boxplot,
	boxplot prepared={draw position=1.5, 
	lower whisker=0.00, 
	lower quartile=0.07, 
	median=0.17, 
	upper quartile=0.27, 
	upper whisker=0.45,
	sample size=1}
	] coordinates{ (1.5, 1.63) (1.5, 0.58) };

\nextgroupplot[
    clip=false,
	boxplot/draw direction=y,
	boxplot/variable width,
	boxplot/every median/.style={black,very thick,solid},
    axis lines=left,
	xmin=0.5,xmax=1.5,
	width=0.33\textwidth,
	height=0.20\textheight,
	ylabel style={align=center}, 
	y tick label style={align=right},
	x tick label style={align=center},
    enlarge y limits=0.05,
	xtick={1},
	ymajorgrids=true,
	grid style=dashed,
	xticklabels={{\small OR-Tools}}, xticklabel style={rotate=0}, ylabel={\footnotesize Average \% gap},
	scatter/classes={ a={mark=star}, b={mark=*}}]

	\addplot[
	mark=*, mark size=0.5pt,boxplot,
	boxplot prepared={
	draw position=1,
	lower whisker=0.38,
	lower quartile=2.835,
	median=3.945,
	upper quartile=4.965,
	upper whisker=8.040,
	sample size=1
	}
	] coordinates{(2, 8.62)};

\end{groupplot}

\end{tikzpicture}